\documentclass{article}
\usepackage{amsmath,epsfig}
\usepackage{amssymb}
\usepackage[preprint]{spconfa4}

\let\OLDthebibliography\thebibliography
\renewcommand\thebibliography[1]{
  \OLDthebibliography{#1}
  \setlength{\parskip}{0pt}
  \setlength{\itemsep}{0pt plus 0.3ex}
}

\usepackage[backend=biber, style=ieee, doi=false,isbn=false,url=false,eprint=false]{biblatex}
\begin{document}\sloppy

\title{ICME 2022 Few-shot LOGO detection top 9 solution}
%
\name{Ka Ho Tong$^{\ast}$, Ka Wai Cheung$^{\ast}$ and Xiaochuan Yu$^{\ast}$}
\address{$^{\ast}$Wisers AI Lab \\ hillyu@wisers.com}

\maketitle

\begin{abstract}
ICME-2022 few-shot logo detection competition is held in May, 2022. Participants are required to develop a single model to detect logos by handling tiny logo instances, similar brands, and adversarial images at the same time, with limited annotations. Our team achieved rank 16 and 11 in the first and second round of the competition respectively, with a final rank of 9th. This technical report summarized our major techniques used in this competitions, and potential improvement.
\end{abstract}
\section{Introduction}

A logo detection model can be a valuable product, especially in e-commerce industry where platforms are struggling to protect their customers from malign usage and plagiarism of their brand logos. However, labeled data is usually expensive and sometimes hard to acquire in abundance. To simulate the typical scenario in the business, this challenge requires participants to train a model which can detect tiny and similar logo instances, while overcoming the difficulties introduced by limited labeled data. This competition consists of 2 stages. In the first stage, the model is required to detect 50 categories, with 50 labeled images for each category. In the final stage, there are another 50 categories, with only 20 labeled images for each category.

In this paper, we briefly summarize our techniques used throughout the competition. We will focus on methods that are proven particularly effective in this competition, while omitting some common approaches that are widely available and well documented somewhere else. Below is a brief summary of what to be covered in this report:
\begin{itemize}
\item Baseline and hardware setup.
\item Augmentations and corresponding ablation studies conducted to understand their effectiveness in this competition.
\item Model fine-tuning strategy and post-processing methods that are proven significantly effective to the final mAP.
\end{itemize}

\section{Baseline}
We used Cascade-RCNN \cite{cai_cascade_2018} as our base-architecture in every stage of the competition, as it performs well in the major object detection tasks such as COCO object detection. Because of the hardware limitation, we used a middle-size backbone such as Resnet101 or Res2net101 \cite{gao_res2net_2021}. It turned out both architectures performed equally well in both rounds of the competition. Inspired by the strong baseline mentioned in Ghiasi, \textit{et.al.}\cite{ghiasi_simple_2021}, we adopted a strong scale-jittering with random horizontal flipping as our basic augmentation pipeline. We left out 5 images for each class from the training dataset to form a validation set, which was later used for hyperparameter tuning and ablation study. Those images were added back in the final training for the submission. All of our experiments are conducted on a hardware with 3 x V100 GPUs, and each round of experiments took less than 24 hours to finish. We used open-source framework `mmdetection' \cite{chen_mmdetection_2019} during our model training and experiments. The following is our baseline setting:
\begin{itemize}
\item Training Image resolution: 1200 x 1200
\item Total batch Size: 15
\item Optimizer: AdamW with learning rate 0.0001 with L2 weight decay = 0.1
\item Training Epochs: 150 epochs with learning rate decay by 0.1 at 132 and 144 epoch.
\end{itemize}
Our baseline achieved 65.8 mAP in first-round validation set and 56 mAP in the testing set used by the leaderboard. 

\section{Augmentations}

An effective way to increase the diversity of training images is performing image augmentations. We found that this method is very effective in handling few-shot object detection problem, and it contributed most to our mAP improvement. The following subsections lists the most effective augmentations we adopted in this competition.

\subsection{Random rotation}

Based on the assumption that logos are shape invariant under rotations, we performed an aggressive rotation on each image. Each image has a probability to be rotated by 90*n (n=1,2,3) degrees clock-wisely. 

\subsection{Simple-Mixup}

Inspired by the simple copy-paste augmentation by Ghiasi, \textit{et.al.} \cite{ghiasi_simple_2021} used in instance segmentation, we designed and implemented a so-called `Simple-Mixup' by combining strong scale-jittering and mixup together. At each iteration, two images are randomly picked and scale-jittered by a ratio selected from 0.1 to 2.0 separately. Two images are then padded into the same shape and blended with alpha = 0.5.

\subsection{Strong Color Jittering}

Based on another assumption that logos are usually determined by the shape instead of the color, we performed strong color jittering on each image. It includes a sequence of operations, such as 1. randomly invert the image for all three color channels; 2. randomly change its brightness, contrast, saturation and hue value; 3. randomly swap the color channels; 4. randomly apply Gaussian blur, Gaussian noise and impulse noise. 

\subsection{RandAugment}

We additionally implemented and applied RandAug \cite{cubuk_randaugment_2020} after the color jittering. We did a grid search and found that (N=1, M=10) is best setting for this dataset. For COCO, the best setting is (N=1, M=5), the difference is very likely due to the difference of detection difficulty between these two datasets.

\subsection{Ablation study}
 
Using above four extra augmentations, we obtained 70.5 mAP in the validation set in the round 1, and roughly 65 mAP on the leader-board. (We implemented RandAug in the second round of the competition, so the expected result should be slightly higher). Table \ref{tab:1} shows the ablation study of the data augmentation. It shows that strong color jittering produced most significant mAP gain.  

\begin{table}[t]
\begin{center}
\caption{Ablation study on data augmentation. ROT stands for random rotate by 90 degrees; MIX stands for simple Mixup; SCJ stands for strong color jittering and RA stands for RandAugment. Simple Mixup is replaced by strong scale-jittering in case of absence. } \label{tab:1}
\begin{tabular}{|c|c|c|c|c|c|}
  \hline
  ROT & MIX & SCJ & RA & mAP & $\Delta$ mAP \\
  \hline
  X & $\checkmark$ & $\checkmark$ & $\checkmark$ & 69.5 & -1.0 \\
  \hline
  $\checkmark$ & X & $\checkmark$ & $\checkmark$ & 68.5 & -2.0 \\
  \hline
  $\checkmark$ & $\checkmark$ & X & $\checkmark$ & 68.2 & -2.3 \\
  \hline
  $\checkmark$ & $\checkmark$ & $\checkmark$ & X & 69.2 & -1.3 \\
  \hline
  $\checkmark$ & $\checkmark$ & $\checkmark$ & $\checkmark$ & 70.5 & 0 \\
  \hline
\end{tabular}
\end{center}
\end{table}

\section{Other modifications}

Other effective modifications were also adopted in the first and second round of the competition. We changed the number of convolution layers to 2 layers in RPN heads, which leads 0.5 mAP gain; changing the loss function of the regression head from smooth L1 loss to balanced L1 loss \cite{pang_libra_2019} gives 0.6 mAP gain. Increasing the training and testing resolution from 1200x1200 to 1472x1472 gives 0.9 mAP gain roughly. Testing time augmentation with multi resolution [982x982, 1472x1472, 2208x2208] and randomly horizontal flipping gives 1.0 mAP gains. Doubling the pre/post number of proposals in the RPN head gives 0.2 mAP gain. Our final model on round 1 is 73.0 mAP on validation set and 66.6 mAP on the testing set.

For the second round, we used model weights from first round as pretrain. we used the same setting as first round with two exceptions, which gives roughly 0.3 mAP gain: 
\begin{itemize}
\item The backbone's learning rate is scaled by 0.1. 
\item The cls-head and reg-head in RCNN is re-initialized.
\end{itemize} 

We also used an extra post-processing method by rescaling the score of detected bounding boxes. We found that in the training set, most of the images have single class of annotated logo. Thus we assumed the same observation holds for test set. At inference stage, for a given image the bounding box with highest class score is treated as the `major class'. The detected bounding boxes not belonging to the `major class` are suppressed by scaling with a constant factor.  This gives us 1.1 mAP gain. 

Finally, noticing that there are 2 common brands appeared in both the first round and the second round, we added 100 labeled images from the first round (50 images for each brand) as additional training data when training the second round model. The increment of training data gives an extra 0.7 mAP gain. For the second round, our final model is 59.6 mAP on the validation set and 61.6 mAP on the testing set.

\printbibliography
\end{document}